\documentclass{article}
\usepackage{pifont}

\usepackage{ijcai26}

\usepackage{times}
\usepackage{soul}
\usepackage{url}
\usepackage[hidelinks]{hyperref}
\usepackage[utf8]{inputenc}
\usepackage[small]{caption}
\usepackage{graphicx}
\usepackage{amsmath}
\usepackage{amsthm}
\usepackage{booktabs}
\usepackage{algorithm}
\usepackage{algorithmic}
\usepackage[switch]{lineno}
\usepackage{xcolor}
\usepackage{colortbl}
\usepackage{verbatim}
\usepackage{pifont}
\usepackage{makecell}
\usepackage{booktabs}
\usepackage{colortbl}

\newcommand{\cmark}{\ding{51}}  % ✓
\newcommand{\xmark}{\ding{55}}  % ✗
\definecolor{lightred}{RGB}{255,210,210}
\definecolor{lightblue}{RGB}{242,238,255}
\definecolor{red}{RGB}{220,110,110}
\definecolor{purple}{RGB}{160,130,215}
\title{UniSync: Towards Generalizable and High-Fidelity Lip Synchronization for Challenging Scenarios}

\author{
    Ruidi Fan$^1$,
    Yang Zhou$^1$, 
    Siyuan Wang$^1$\thanks{Corresponding Author},
    Tian Yu$^1$,
    Yutong Jiang$^1$,
    Xusheng Liu$^1$
    \affiliations
    $^1$Mango TV
    \emails
    \{fanruidi, zhouyang, yutian6, jiangyutong, liuxusheng\}@mgtv.com, syyxsxx411@gmail.com
}

\begin{document}

\maketitle

\begin{abstract}
    Lip synchronization aims to generate realistic talking videos that match given audio, which is essential for high-quality video dubbing. However, current methods have fundamental drawbacks: mask-based approaches suffer from local color discrepancies, while mask-free methods struggle with global background texture misalignment. Furthermore, most methods struggle with diverse real-world scenarios such as stylized avatars, face occlusion, and extreme lighting conditions. In this paper, we propose UniSync, a unified framework designed for achieving high-fidelity lip synchronization in diverse scenarios. Specifically, UniSync uses a mask-free pose-anchored training strategy to keep head motion and eliminate synthesis color artifacts, while employing mask-based blending consistent inference to ensure structural precision and smooth blending. Notably, fine-tuning on compact but diverse videos empowers our model with exceptional domain adaptability, handling complex corner cases effectively. We also introduce the RealWorld-LipSync benchmark to evaluate models under real-world demands, which covers diverse application scenarios including both human faces and stylized avatars. Extensive experiments demonstrate that UniSync significantly outperforms state-of-the-art methods, advancing the field towards truly generalizable and production-ready lip synchronization.
\end{abstract}

\begin{figure}[t]
    \centering
    \includegraphics[width=\columnwidth]{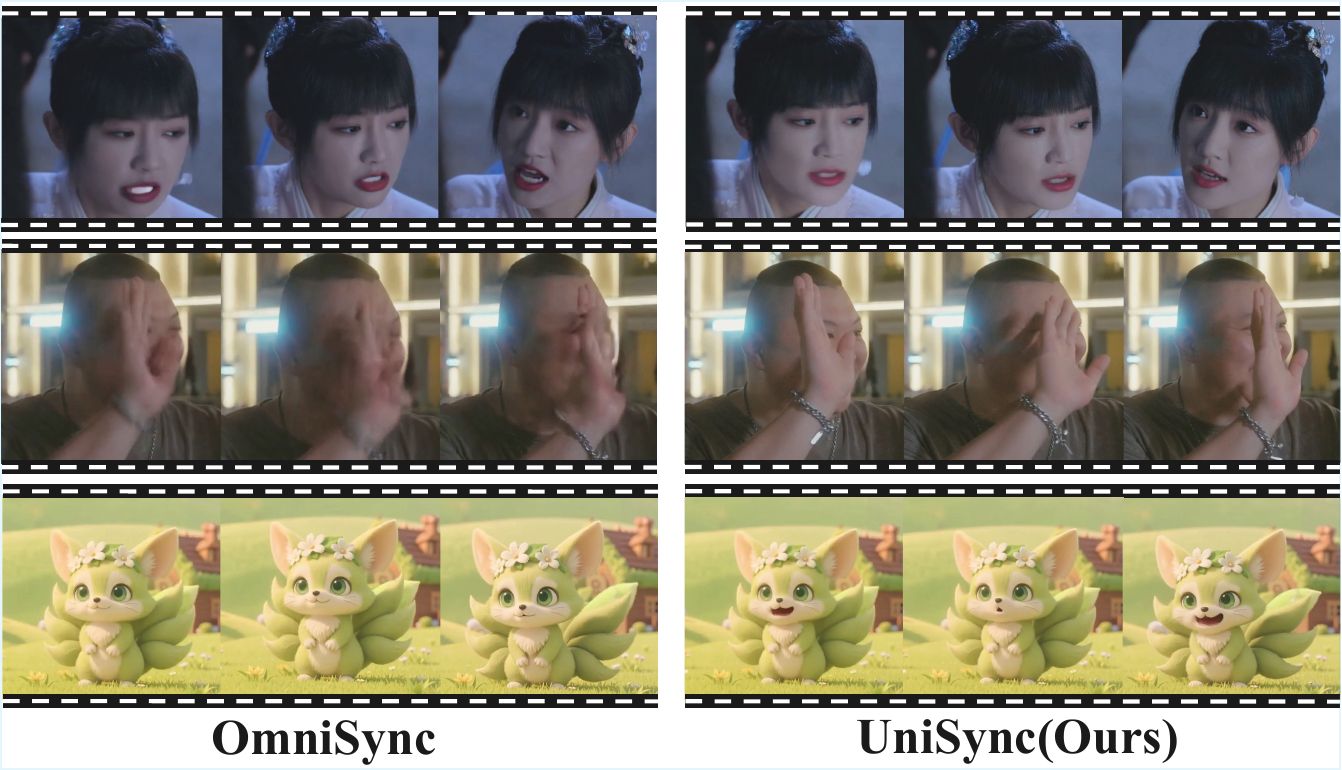}
    %\caption{The comparison of lip synchronization results in real-world scenarios between previous SOTA method OmniSync and our proposed method UniSync. Our model perform a better result under low-light condition, facial occlusion and anthropomorphic cartoon, while maintaining high visual quality and temporal coherence.}
    \caption{Comparison with OmniSync on challenging real-world scenarios. While OmniSync shows degraded quality under low-light conditions and occlusions, it completely fails on stylized cartoons, producing no lip modifications. Our method robustly handles all cases with accurate synchronization and high visual fidelity.}
    \label{fig:abs}
\end{figure}

\section{Introduction}
Digital video content grows rapidly on entertainment, education, and social media platforms. Lip synchronization, the task of synchronizing lip movements with input audio, has become essential for multilingual localization and video content production.

For lip synchronization, two main paradigms have been identified. Reference-driven portrait animation approaches~\cite{styletalk,vexpress,omnihuman,omniavatar,wans2v,infinityhuman,fantasytalking} create talking portraits from static images by using audio-conditioned diffusion. While maintaining background and character motion, video dubbing approaches~\cite{diff2lip,musetalk,latentsync,omnisync,sayanything,syncanyone} modify the mouth area in pre-existing videos. This paper focuses on the video dubbing task.

Current video dubbing methods often struggle to keep local editing quality and global consistency, facing a significant challenge in generating stable and high-fidelity videos. In general, these methods can be categorized into two paradigms: mask-based methods, which strictly confine editing to a masked mouth region, and mask-free methods, which process the full video frame to adaptively synthesize lip motion.% In general, these methods can be categorized into two paradigms: mask-based methods, which only permit editing in a masked mouth area, and mask-free methods, which process the entire video frame to adaptively create lip motion.

Mask-based approaches restrict generation to a fixed mouth mask to preserve the background, but this introduces a challenging blending problem. The generated lip area often fails to match the original lighting and skin texture, which leads to visible seams or color shifts at mask boundaries and influences the visual fidelity. Moreover, strict masking limits natural jaw motion, making speech appear stiff. In contrast, mask-free methods generate the entire frame to enable smoother color transitions. But they will have unintended modifications in non-target areas such as hair, facial contours, or background details by removing spatial constraints, leading to lose anchoring with the original video. Thus, the generated videos no longer satisfy the precise requirements of video dubbing. Furthermore, most existing models are trained on open-source talking-head datasets featuring fixed cameras, uniform lighting, and close-up shots. This makes them perform much worse on real-world content. These methods can not perform well in professional production workflow because it could not precisely find out the location of the mouths or made the texture in the movies were consistent with strong shadows, different viewing angles, different styles of avatar.

To this end, we propose UniSync, a unified lip synchronization framework for different scenarios. Our method fine-tunes a pretrained audio-image-to-video (AI2V) model with a small training dataset and obtains excellent results through a good design framework. In the training process, we propose a mask-free pose-anchored approach to avoid color artifacts due to hard masks. Instead of masking the mouth, we input the full image and use pose data to anchor the facial structure, forcing the model to learn natural lip animation while maintaining head stability. This is to remove the reliance on good masks and to handle large head movements. Although the training stage guarantees the quality of generation, inference requires maintaining strict consistency in non-modified areas. We introduce a mask-based blending consistent inference mechanism that operates in both latent and pixel spaces. In the latent space, Temporal-Adaptive Latent Injection (TALI) fuses the noisy latents of the original video into the generation process to suppress texture shifts and identity drift. In the pixel space, a Gaussian-based smooth compositing strategy is used to ensure the smooth transition of the spatial space after the edited lips and the original face, and avoid boundary discontinuities. %This dual-space approach successfully solves the generation problem of the mask-free method before this paper.

Data diversity in addition to the model is necessary for generalization. We extract a small set of 5000 videos including movies, TV series and cartoons, and then we fine-tune the model with LoRA, which can achieve a good generalization effect at a low training cost. A targeted fine-tune plan is beneficial for domain adaptation and will probably work in messier situations, such as stylized avatars, face occlusion, and extreme lighting. We also introduce the RealWorld-LipSync benchmark, which contains difficult cases, including anthropomorphic cartoons and extreme lighting, for the test of the model's performance under real conditions rather than in an ideal environment.

The main contributions are summarized as follows:
\begin{itemize}
    \item We propose UniSync, a unified lip synchronization framework that eliminates color artifacts and maintains head motion by means of mask-free pose-anchored training to achieve high local and global consistency.
    \item We introduce a novel mask-based blending mechanism when inference, including two key strategies: temporal-adaptive latent injection and Gaussian-based smooth compositing, making sure that the consistency of the unmodified part and the natural blending of the background.
    \item We build the RealWorld-LipSync benchmark to measure the performance of dubbing on various real-world scenarios, to link research to the real world.
\end{itemize}

\section{Related Works}

\subsection{Video Diffusion Models}
%Diffusion models have significantly facilitated video generation. Video diffusion models such as SVD~\cite{svd}  and DiT~\cite{dit} have established strong baselines by extending 2D architectures with temporal modeling or adopting pure transformer designs. Recent large-scale models like Sora~\cite{sora} and Self-Forcing++~\cite{selfforcing++} further extend these capabilities to multi-minute  generation.  While these models offer powerful generative priors, directly applying them to localized editing tasks like lip synchronization remains challenging, as standard diffusion processes often struggle to modify specific facial regions without altering the surrounding visual integrity.
Diffusion models have been made progress in generation videos as well as DiT~\cite{dit} provides the scalable backbones for generating the high quality videos. Video diffusion models have developed strong baselines by either extending 2D architecture with time layer~\cite{vdm,svd} or using pure transformer design~\cite{wan2.1,latte}. Recent large scale models ~\cite{freenoise,freepca,DiTCtrl,streamingt2v,fifo,cono,sora,selfforcing++,hunyuanvideoavatar,skyreelsv2} further extend video generation capactiy to over tens of seconds. Although these models have powerful generative priors, they are difficult to use directly for localized editing tasks such as lip synchronization, because standard diffusion processes tend to be have poor results when making changes in a certain part of the face without causing damage to the visual completeness of the rest of the face.
%Diffusion models have significantly improved video generation, with DiT~\cite{dit} provides a scalable backbone for high-quality synthesis. Video diffusion models have established strong baselines by either extending 2D architectures with temporal layers~\cite{vdm,svd} or adopting pure transformer designs~\cite{wan2.1,latte}. Recent large-scale models ~\cite{freenoise,freepca,DiTCtrl,streamingt2v,fifo,cono,sora,selfforcing++,hunyuanvideoavatar,skyreelsv2} further extend video generation capability with more than tens seconds. Although these models have powerful generative priors, directly applying them to localized editing tasks like lip synchronization remains challenging, as standard diffusion processes often struggle to modify specific facial regions without altering the surrounding visual integrity.

\begin{figure*}[t]
\centering

    \includegraphics[width=0.9\linewidth, height=0.35\textheight, keepaspectratio]{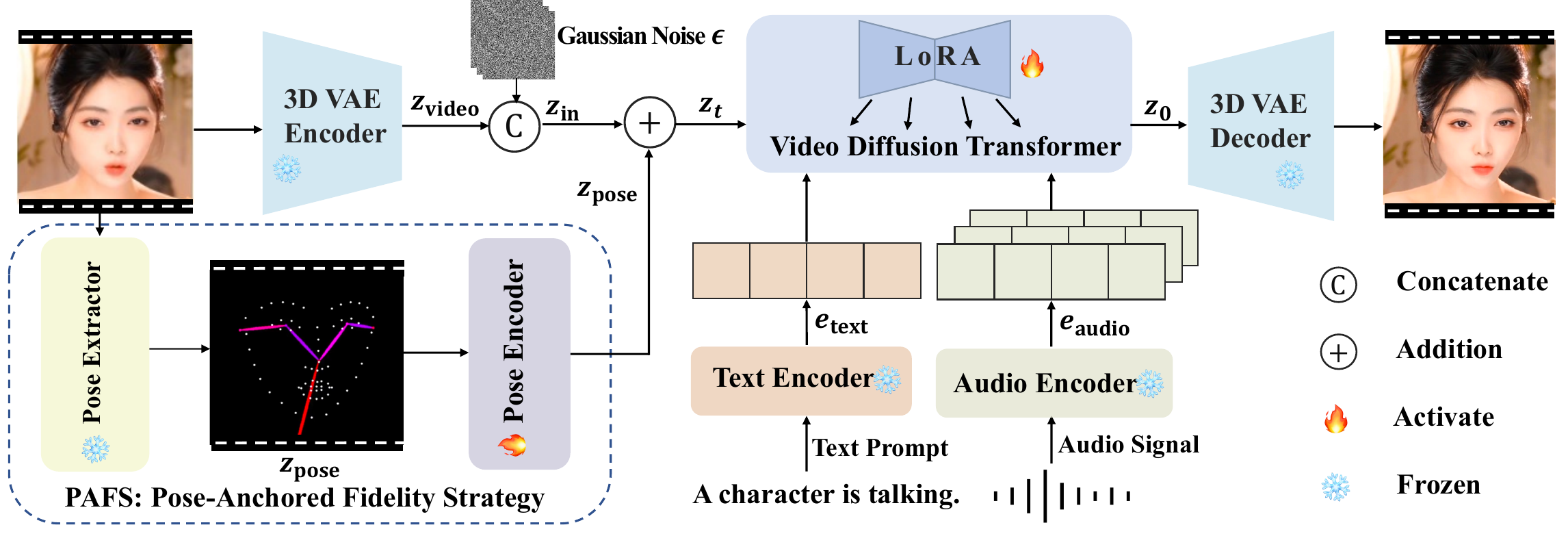}
    \caption{Training framework of the proposed method. The Pose-Anchored Fidelity Strategy (PAFS) enforces a direct mapping between pose variation and facial motion. LoRA-based fine-tuning makes the model better adapt to pose-anchored, mask-free training efficiently.}
    \label{fig:train}

\vspace{1em}

    \includegraphics[width=1\linewidth, keepaspectratio]{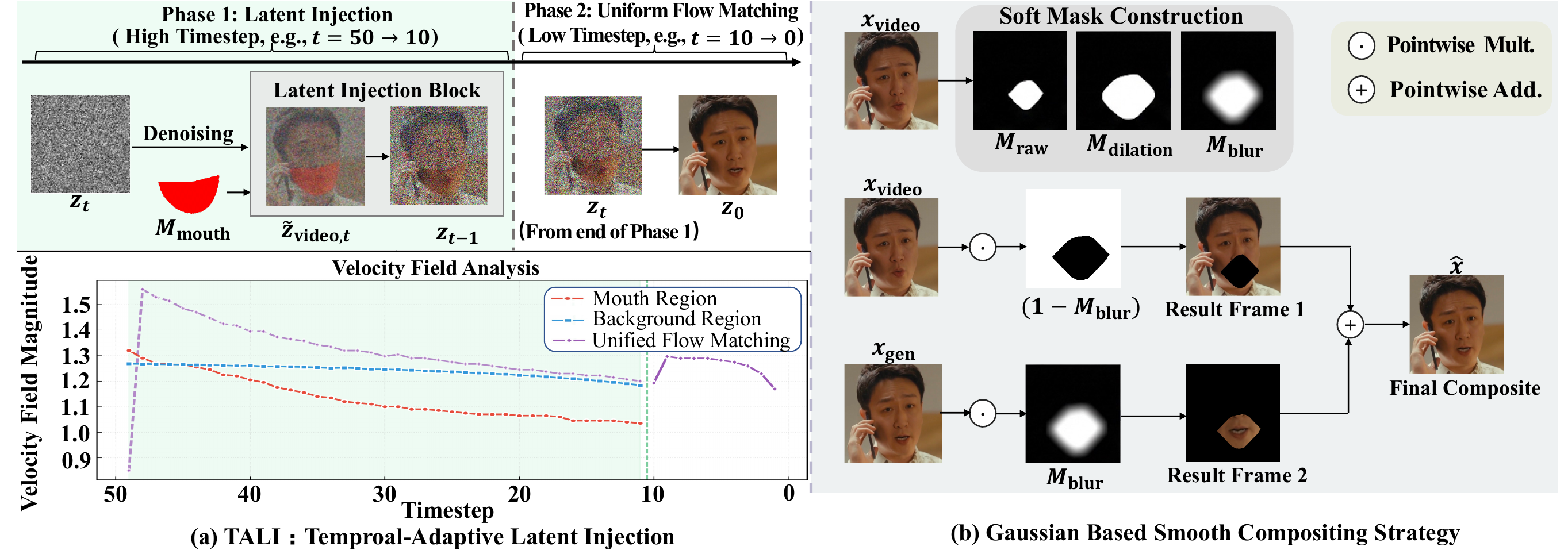}
    \caption{Overview of the inference framework. (a) Illustration of TALI. At high timesteps, only mouth-region denoising latents are preserved while non-mouth regions use ground truth latents with timestep-dependent noise for background consistency. At low timesteps, uniform flow matching ensures global distribution consistency. The velocity field analysis shows decreasing magnitude in the mouth region during latent injection, while background remains stable, followed by unified denoising in the flow matching phase. (b) Gaussian Based Smooth Compositing. The mouth mask $M_{\text{raw}}$ is processed via Gaussian dilation ($M_{\text{dilation}}$) and blurring to produce the soft mask $M_{\text{blur}}$. The generated $x_\text{gen}$ is then composited with ground truth $x_\text{video}$ using $M_{\text{blur}}$ to yield the final output $\hat{x}$.}
    \label{fig:infer}
\vspace{-10pt}
\end{figure*}

\subsection{Audio-driven Lip Synchronization}
According to the usage scenarios of audio-driven lip synchronization, it can be categorized as portrait animation and video dubbing. Portrait animation methods~\cite{sadtalker,dreamtalk,difftalk,emo,visa1,hallo} make the talking head out of static images, and it can reach very high visual quality through full-frame reconstruction. However, these are design plans for content production, not content editing, thus unusable for the video dubbing described in this paper which needs to preserve the background, lighting, and scene context of the video.
While video dubbing, on the other hand, involves making fine edits to the mouth region of the video, while leaving all other visual components as they are. Earlier works such as Wav2Lip~\cite{wav2lip}, VideoReTalking~\cite{videoretalking} use GANs for synchronization but often have blurred textures. Although recent Diffusion-based methods like Diff2Lip~\cite{diff2lip} and DINet~\cite{dinet} improve visual quality, the majority of methods still have problems: even though mask-based methods such as LatentSync~\cite{latentsync} and MuseTalk~\cite{musetalk} are able to retain the background, but it also causes color artifacts due to lighting and texture mismatches. On the contrary, mask-free approaches (OmniSync~\cite{omnisync}) can circumvent hard boundaries of face mask but run into troubles of large-scale texture artifacts and uncontrolled changes in non-target region. So we introduce a novel framework named UniSync to solve above problems.

\section{Method}

\subsection{Overview}
\textbf{Task.} 
Video dubbing aims to synchronize the visual lip movement and different audio. Given a source video and an audio, our aim is to make local changes to the mouth area so as to make the lip movements match the input audio accurately and keep the original head pose, facial identity, background, and all the dynamics of the video as well. So we introduce a mask-free training and mask-based inference pipeline with targeted strategies for this.

\textbf{Pipeline.}
As shown in Figure~\ref{fig:train}, we introduce Pose-Anchored Fidelity Strategy (PAFS) to anchor global head structure and regulate motion in the diffusion latent space during the training. On the basis of recent audio-image-to-video (AI2V) models which have good lip synchronization, we adapt a pre-trained model to the video dubbing task through LoRA fine-tuning on a small but diverse dataset. This adaptation can achieve stable and localized editing of lips in the face, and head motion information can be kept, and it also has higher adaptability for different types of videos. In addition, mask-free training reduces the color differences of prior mask-based methods.

An illustration of inference is given by Figure~\ref{fig:infer}, we first introduce the Temporal-Adaptive Latent Injection (TALI), the two-stage denoising strategy. At higher timesteps, only the mouth region does the flow-matching denoising, and the non-mouth regions are constrained by timestep-dependent noisy ground truth latents to ensure background and identity consistency. At low time steps, uniform flow matching denoising is carried out to maintain global distribution consistency. After that, we use a Gaussian-based smooth composite strategy to back-integrate the editing of the mouth region into the original video in the pixel space to get a realistic effect.

%Inference is illustrated in Figure~\ref{fig:infer}, we firstly introduce Temporally Adaptive Latent Injection(TALI), a two-stage denoising strategy. At high timesteps, only the mouth region do the flow-matching denoising, while constraining non-mouth regions with timestep-dependent noisy ground-truth latents to enforce background and identity consistency. At low timesteps, uniform flow matching denoising is applied to ensure global distribution coherence. After that, a Gaussian-based smooth compositing strategy seamlessly integrates the edited mouth region back into the original video in the pixel space, producing temporally stable and photorealistic results.

\subsection{Mask-Free Pose-Anchored Training}
We propose a pose-anchored Diffusion Transformer for high-fidelity video dubbing. In contrast with the mask-based video dubbing methods like LatentSync ~\cite{latentsync} which have strict requirements for the mouth mask, if there is not a powerful enough face detector, it will fail at the task. We choose the mask-free training method to take the whole face video as the input and output the whole video. So that we can solve the color inconsistencies between the edited mouth and the original video. And we also introduce an effective module named Pose-Anchored Fidelity Strategy (PAFS), which anchors head motion to the explicit pose variations. Then the model learns robust motion variations even when the head is moving complicatedly and also doesn’t have to use hard masks to isolate the mouth region. This mask-free pose-anchoring training strategy can reduce the common color deviation and structural artifact problems in mask-based lip editing methods.

\subsubsection{Video Latent Concatenation for AV2V Training}
We first modify the pretrained audio-image-to-video (AI2V) model into an audio-video-to-video (AV2V) model. Taking $V_{\text{in}}\in{R^{F\times H\times W\times 3}}$ as the input video, we get:
\begin{equation}
    z_\text{video} = \text{VAEenc}(V_{\text{in}})\in R^{B\times C\times f \times h \times w}.
\end{equation}
Where $F$ refers to the number of frames, $H$, $W$ are spatial dimensions, $B$ is the batch size, $C$ denotes the pose latent channel dimension, $f$ denotes the compressed frame dimension and $h, w$ are the down-sampled spatial dimensions in the latent space. As for the training of the AV2V model, the video latent $z_\text{video}$ is concatenated to the original latent $\epsilon \sim \mathcal{N}(0, 1)$ to build the model input:
\begin{equation}
    z = \text{concat}_{\text{channel}}(\epsilon, z_{\text{video}}) \in R ^{B\times 2C\times f \times h \times w}.
\end{equation}

Our model is trained with a flow matching objective that learns the continuous transport from noise to data distributions. The velocity field is defined as $v = \epsilon - z_\text{video}$, and the velocity prediction network $\hat{v}_\theta$ takes the concatenation of the input $z$ along with conditioning signals $c_{\text{aud}}$, $c_{\text{pose}}$, and $c_{\text{text}}$. Training loss is:
\begin{equation}
\resizebox{0.9\linewidth}{!}{
$
\begin{aligned}
    \mathcal{L}_{\mathrm{MSE}} = E_{\textstyle \epsilon, z_{\text{video}}, t, c }[ \omega(t) \cdot \| v - \hat{v}_\theta(z, t, c_{\text{aud}}, c_{\text{pose}}, c_{\text{text}}) \|_2^2].
\end{aligned}
$
}
\end{equation}
Where $t$ is the diffusion timestep, and $\omega(t)$ is a timestep-dependent weighting function that gives more weight to the intermediate steps which are important for learning smooth motion dynamics. Minimize this objective to make the model learn how to synthesize video frames according to the original input video, rather than just rely on the reference frame to synthesize the video.

However, training only the AV2V model is inadequate. We observe that the generated head motion still meets the drift problem, which makes it hard to put the composite new face seamlessly back onto the original background. We thus advocate our Pose-Anchored Fidelity Strategy and bring in the explicit motion anchoring to enforce this strict correspondence.

\subsubsection{Pose-Anchored Fidelity Strategy}
Pose-Anchored Fidelity Strategy (PAFS) enforces a direct mapping from pose variations to facial motion. As shown in Figure~\ref{fig:train}, PAFS extracts the explicit pose information from the input video and injects it as a structural anchor into the diffusion process through additive latent fusion. This strategy can maintain the temporal continuity and spatial consistency of various head movements from the model without region-specific masks.

Given a face video as input, we first use a pretrained pose extractor to obtain the pose keypoints of each frame. Using RTMPose~\cite{rtmpose} as the pose extractor, it can give stable head pose estimation with face landmarks and head direction information. These keypoints are rendered as a pose video $V_{\text{pose}}\in{R^{F\times H\times W\times 3}}$, in which $F$, $H, W$ are the same dimensions with $V_{\text{in}}$. The pose video is then encoded into the latent representation using the same VAE encoder with input video,
\begin{equation}
    z_\text{pose} = \text{VAEenc}(V_{\text{pose}})\in R^{B\times C\times f \times h \times w}.
\end{equation}
Where $B$, $C$, $f$, $h$ and $w$ are the same dimensions with $z_\text{video}$.

To inject pose information into the diffusion transformer at the feature level, we design a pose encoder that progressively transforms the pose latent $z_{\text{pose}}$ into pose-anchored tokens matching the shape of video tokens $z_{in}$. The pose encoder consists of two stages:
We first use a 3D convolutional layer to extract the pose features at the patch level and reduce the spatial-temporal resolution to be consistent with the space of the video tokens.
\begin{equation}
    z_{\text{pose}}^{(1)}={\text{Conv3D}}_{k,s}(z_\text{pose}) \in R^{D \times {f'} \times {h'} \times{w'}}.
\end{equation}
Where $\text{Conv3D}_{k,s}$ is a 3D convolution with kernel size $k$ and stride $s$(both set to the patch size), $D$ is the target feature dimension, and $f^{'}, h^{'}, w^{'}$ are the down-sampled temporal and spatial dimensions.
Patch features are flattened into a token sequence and projected via learnable linear transformation followed by layer normalization.
\begin{equation}
    z_{\text{pose}}^{(2)} = \text{Linear}(\text{Flatten}(z_{\text{pose}}^{(1)})) \in R^{S \times D},
\end{equation}
\begin{equation}
    z_{\text{pose}} = \text{LayerNorm}(z_{\text{pose}}^{(2)} + \text{PE}) \in R^{B \times S \times D}.
\end{equation}
Where $S = f' \times h' \times w'$ refers to the number of tokens, PE represents the learnable positional embedding, and $B$ represents the batch size. This hierarchical design can get both pose encoder local pose detail and global pose pattern by convolution and linear projection respectively.

The core composition of PAFS is the additive fusion of pose features with video features. Unlike concatenation-based conditioning that treats pose as additional information, our additive fusion treats pose as a persistent structural bias throughout the diffusion process. The concatenated video latent is embedded into tokens through the patch embedding layer, matching the demension of $z_\text{pose}$.
\begin{equation}
    z_\text{in} = \text{PatchEmbed}(z) \in R ^{B\times S \times D}.
\end{equation}
The final pose-anchored input for the Diffusion Transformer is obtained by element-wise addition:
\begin{equation}
    z_t = z_\text{pose} + z_{\text{in}}.
\end{equation}

\subsection{Mask-Based Blending Consistent Inference}
Although our mask-free pose-anchored training can solve the color artifacts problem, if we directly use the mask-free inference, it may bring unintended changes to the face or background area, causing visual inconsistencies. To this end, we propose a mask-based blending inference pipeline that selectively retains the original information on non-speaking regions and injects the generated lip motion through Temporal-Adaptive Latent Injection and Gaussian-based Smooth Compositing Strategy.

\subsubsection{Temporal-Adaptive Latent Injection}
Temporal-Adaptive Latent Injection (TALI) aims to avoid the problem of texture inconsistency and identity deviation in non-mouth area during diffusion inference. The key insight is that diffusion models are more sensitive to the overall structure and appearance of the image at the early stage of denoising (high-noise timesteps), and then later steps make fine-grained modifications to the local details. Therefore, rather than constrain the entire process, we directly inject ground truth video latents into non-mouth regions only at high-noise timesteps to maintain global consistency and allow the model to generate lip motion freely.

We define an injection ratio $\tau_{\text{inj}} \in [0,1]$ for the range of the time-domain injected ground truth. The number of steps is $T$, indexed as $t=T$ (pure noise) to $t=1$ (clean output). At each timestep $t$, it predicts the velocity $\hat{v}$ and updates the latent via flow matching: $z_{t-1} \leftarrow z_t - \Delta t \cdot \hat{v}$. When $t > (1 - \tau_{\text{inj}}) \cdot T$ (early high-noise period), we replace the non-mouth area of $z_{t-1}$ with properly noised ground truth latents:

\begin{equation}
z_{t-1} = M \odot z_{t-1} + (1 - M) \odot \tilde{z}_t.
\end{equation}

Where $M \in {0, 1}^{B\times 1\times f \times h \times w}$ is a binary mouth mask in latent space (value 1 in mouth region, 0 elsewhere), and $\tilde{z}_t$ is the noised ground truth latent:
\begin{equation}
\tilde{z}_t = \alpha_t z_{\text{video}} + \sigma_t \epsilon', \quad \epsilon' \sim \mathcal{N}(0, \mathbf{I}).
\end{equation}

Where $\alpha_t$ and $\sigma_t$ are noise schedule parameters satisfying $\alpha_t^2 + \sigma_t^2 = 1$ ensuring $\tilde{z}_t$ matches the noise level of $z_t$ at timestep $t$.

\begin{algorithm}[tb]
\caption{TALI: Temporal-Adaptive Latent Injection}
\label{alg:tali}
\textbf{Input}: Video latent $z_{\text{video}} \in R^{B \times C \times f \times h \times w}$, mouth mask $M \in \{0,1\}^{B \times 1 \times f \times h \times w}$, conditions $c_{\text{aud}}, c_{\text{pose}}, c_{\text{text}}$\\
\textbf{Parameter}: Injection ratio $\tau_{\text{inj}} \in [0,1]$, denoising steps $T$\\
\textbf{Output}: Edited video latent $z_0$
\begin{algorithmic}[1]
\STATE Sample initial noise $z_T \sim \mathcal{N}(0, \mathbf{I})$
\FOR{$t = T$ \TO $1$}
    \STATE $\hat{v} \leftarrow \text{Model}([z_t; z_{\text{video}}], t, c_{\text{aud}}, c_{\text{pose}}, c_{\text{text}})$
    \STATE $z_{t-1} \leftarrow z_t - \Delta t \cdot \hat{v}$ \hfill \COMMENT{Flow matching update}
    \IF{$t > (1 - \tau_{\text{inj}}) \cdot T$}
        \STATE $\tilde{z}_t \leftarrow \alpha_t z_{\text{video}} + \sigma_t \epsilon'$, $\epsilon' \sim \mathcal{N}(0, \mathbf{I})$
        \STATE $z_{t-1} \leftarrow M \odot z_{t-1} + (1 - M) \odot \tilde{z}_t$ \hfill \COMMENT{Inject GT}
    \ENDIF
\ENDFOR
\STATE \textbf{return} $z_0$
\end{algorithmic}
\end{algorithm}

As shown in Algorithm~\ref{alg:tali}, in the early denoising stage, the model's prediction in non-mouth regions is replaced by $\tilde{z}_t$, thus maintaining the original texture and identity, and the mouth region is completely fully synthesized under the audio condition. Once $t \leq (1 - \tau_{\text{inj}}) \cdot T$, injection is disabled and the model refines all regions freely, allowing for seamless blending at the boundaries of the regions.

\begin{table}[t]
    \centering
    \small
    \setlength{\tabcolsep}{4pt}
    \begin{tabular}{lccc}
        \toprule
        \textbf{Characteristic} & \textbf{HDTF} & \textbf{AIGC-LipSync} & \textbf{Ours} \\
        \midrule
        \rowcolor{gray!15}
        \textbf{Video Source} & Real-world & AI-generated & Real-world \\
        \textbf{Resolution} & 720-1080p & 1080p & 1080p-4K \\
        \midrule
        \rowcolor{gray!15}
        \textbf{Scene Diversity} & Low & Medium & High \\
        \textbf{Camera Angles} & Frontal & Varied & Diverse \\
        \rowcolor{gray!15}
        \textbf{Lighting} & Uniform & Ideal & Extreme \\
        \textbf{Camera Motion} & Minimal & Limited & Dynamic \\
        \rowcolor{gray!15}
        \textbf{Stylized Content} & \xmark & \xmark & \cmark \\
        \midrule
        \textbf{Applicability} & Medium & Low & High \\
        \bottomrule
    \end{tabular}
    \caption{Comparison of lip synchronization benchmarks. Our benchmark emphasizes challenging real-world conditions including extreme lighting, diverse angles, and stylized content.}
    \label{tab:benchmark_comparison}
\end{table}

\begin{table*}[t]
    \centering
    \resizebox{\textwidth}{!}{
    \setlength{\tabcolsep}{2pt}  % 只减小列间距，不改字体大小
    \begin{tabular}{l|ccccccc|cccccccc}
        \toprule
        & \multicolumn{7}{c|}{\textbf{HDTF Dataset}} & \multicolumn{8}{c}{\textbf{RealWorld-LipSync Benchmark}} \\
        \cmidrule(lr){2-8} \cmidrule(lr){9-16}
        \textbf{Methods} & \textbf{FID}$\downarrow$ & \textbf{FVD}$\downarrow$ & \textbf{CSIM}$\uparrow$ & \textbf{NIQE}$\downarrow$ & \textbf{BRISQUE}$\downarrow$ & \textbf{HyperIQA}$\uparrow$ & \textbf{LSE-C}$\uparrow$ & \textbf{FID}$\downarrow$ & \textbf{FVD}$\downarrow$ & \textbf{CSIM}$\uparrow$ & \textbf{NIQE}$\downarrow$ & \textbf{BRISQUE}$\downarrow$ & \textbf{HyperIQA}$\uparrow$ & \textbf{LSE-C}$\uparrow$ & \textbf{GSR}$\uparrow$ \\
        \midrule
        \textbf{TalkLip} & 16.680 & 691.518 & 0.843 & 6.377 & 52.109 & 44.393 & 5.880 & 15.613 & 447.192 & 0.629 & 8.938 & 56.808 & 41.227 & 4.123 & 71.9\% \\
        \textbf{IP-LAP} & 9.512 & 325.691 & 0.809 & 6.533 & 54.402 & 50.086 & 7.260 & 19.458 & 280.014 & 0.848 & 9.202 & 59.115 & 43.140 & 4.686 & 68.9\% \\
        \textbf{Diff2Lip} & 12.079 & 461.341 & 0.869 & 6.261 & 49.361 & 48.869 & 7.140 & 17.631 & 326.519 & 0.790 & 9.359 & 59.479 & 42.398 & 4.318 & 78.0\% \\
        \textbf{MuseTalk} & 8.759 & 231.418 & 0.862 & \cellcolor{lightblue}5.824 & 46.003 & 55.397 & 6.890 & 16.894 & 174.946 & 0.832 & 8.407 & 59.142 & 44.852 & 4.582 & \cellcolor{lightblue}86.5\% \\
        \textbf{LatentSync} & 8.518 & 216.899 & 0.859 & 6.270 & 50.861 & 53.208 & \cellcolor{lightred}\bf{8.050} & 14.039 & \cellcolor{lightblue}145.060 & \cellcolor{lightred}\bf{0.927} & \cellcolor{lightblue}8.106 & 57.215 & 44.061 & 4.943 & 77.8\% \\
        \textbf{OmniSync} & \cellcolor{lightblue}7.855 & \cellcolor{lightblue}199.627 & \cellcolor{lightblue}0.875 & \cellcolor{lightred}\bf{5.481} & \cellcolor{lightblue}37.917 & \cellcolor{lightblue}56.356 & \cellcolor{lightblue}7.309 & \cellcolor{lightblue}13.655 & 169.867 & 0.868 & 8.518 & \cellcolor{lightred}\bf{52.950} & \cellcolor{lightblue}44.984 & \cellcolor{lightblue}4.983 & 69.1\% \\
        \textbf{UniSync} & \cellcolor{lightred}\bf{6.901} & \cellcolor{lightred}\bf{127.092} & \cellcolor{lightred}\bf{0.887} & 6.258 & \cellcolor{lightred}\bf{34.390} & \cellcolor{lightred}\bf{59.361} & 7.047 & \cellcolor{lightred}\bf{12.976} & \cellcolor{lightred}\bf{141.573} & \cellcolor{lightblue}0.869 & \cellcolor{lightred}\bf{8.023} & \cellcolor{lightblue}55.988 & \cellcolor{lightred}\bf{45.818} & \cellcolor{lightred}\bf{5.091} & \cellcolor{lightred}\bf{93.5\%} \\
        \bottomrule
    \end{tabular}
    }
    \caption{Quantitative comparison with previous methods on HDTF and RealWorld-LipSync benchmarks. The best and second-best results are highlighted in \textcolor{red}{Red} and \textcolor{purple}{Purple}.}
    \label{tab:main_results}
\end{table*}

\subsubsection{Gaussian Based Smooth Compositing Strategy}
The final timestep generates latent is decoded to the generated pixel frames $x_{\text{gen}}$. However, directly pasting the mouth area from these frames onto the original video $x_{\text{video}}$ will result in visible outline artifacts and a decline in the realism of the video.

For this we use a Gaussian Based Smooth Compositing Strategy. As shown in Figure~\ref{fig:infer} (b), we first build a raw binary mask $\mathbf{M}_{\text{raw}}$ that tightly surrounding the mouth area. This mask is then spatially expanded to obtain the dilated mask: $\mathbf{M}_{\text{dilation}}$. Lastly, we apply a Gaussian blur and obtain a smooth spatial varying weighting mask $\mathbf{M}_{\text{blur}}$ where the center of the mouth is marked with value 1 and the edge of the mouth mask is marked with value 0. Finally the composed video $\hat{x}$ is calculated as:
\begin{equation}
  \hat{x} = {M}_{\text{blur}} \odot x_{\text{gen}} + (\mathbf{1} - M_{\text{blur}}) \odot x_{\text{video}}.
\end{equation}

This kind of soft compositing method can restrict the modification area to the mouth region and make smooth transitions at the boundary to retain the background and other non-target faces and parts of the face to avoid disconnection and lack of realism.

\section{Experiments}
\subsection{Datasets}
\subsubsection{Training Dataset}
%To enable robust video dubbing across diverse visual domains, we construct a compact, high-quality dataset for LoRA fine-tuning. The dataset includes 5,000 high-resolution video samples across multiple content types: 2D and 3D cartoons, movies, TV series and TV Shows. These samples capture substantial variation in lighting conditions, visual aesthetics, character appearances, and artistic styles. Despite its relatively small size, the dataset's diversity allows our model to generalize effectively from realistic human videos to stylized cartoon content.
In order to achieve good video dubbing performance across different visual domains, a small amount of high-quality data for LoRA fine-tuning is constructed. There are 5,000 high-resolution video samples of various types of content in the dataset, including 2D and 3D cartoons, movies, TV series and TV shows. These samples cover a wide range of variations in lighting, visual aesthetics, character appearances, etc. As it is of a smaller scale but its variety will make it possible for our model to learn generalize with ease from realistic human videos to stylised cartoon content.

\subsubsection{RealWorld-LipSync Benchmark}
%A critical challenge in lip synchronization research is the disconnection between benchmark evaluation and real-world deployment. Existing datasets like AIGC-LipSync~\cite{omnisync} rely on synthetic videos generated by text-to-video models, which lack visual realism, natural motion dynamics, and challenging real-world conditions such as extreme lighting or stylized content. Consequently, models that perform well on these synthetic benchmarks often fail when applied to professional production workflows.
%A critical challenge in lip synchronization research is the disconnection between benchmark evaluation and real-world deployment. Existing datasets like AIGC-LipSync~\cite{omnisync} rely on synthetic videos lacking visual realism and challenging conditions (Table~\ref{tab:benchmark_comparison}). Consequently, models that perform well on these synthetic benchmarks often fail in professional production workflows.
%To address this gap, we introduce \textbf{RealWorld-LipSync}, a comprehensive benchmark emphasizing real-world complexity rather than dataset size. Our benchmark evaluates lip synchronization under authentic production conditions including extreme lighting, diverse camera angles, and stylized content, providing a rigorous testbed for real-world deployment.
A major problem in the research of lip synchronization is the decoupling of benchmark assessment and actual application. Benchmarks such as AIGC-LipSync~\cite{omnisync} existing rely on the use of synthetic videos that lack realism in the visual and difficult conditions (Table~\ref{tab:benchmark_comparison}). As a result, models that can achieve satisfactory results on these synthetic benchmark data cannot meet the needs in professional production workflows.

In order to address this deficiency, RealWorld-LipSync is proposed, a complete benchmark focused on real-world complexity rather than dataset volume. Lip synchronization benchmark measures in actual production conditions, strong lighting, multiple shooting angles and stylized content of its balance, gives a strict environment of real-life application.

\textbf{Dataset Composition.}
The dataset RealWorld-LipSync included 495 decently picked videos equal or large than 1080p, with a duration from 2 to 15 seconds, and has the 3 real production problem dimensions covered well:
(1) Multiple scene settings: Stage performance scene, indoor conversation scene, outdoor interview scene and scene with camera movement, such as movie scene;
(2) Difficult lighting: the harsh shadows of stage spotlights, the dim light of the interior, the bright light of the day and the darkness of the dark environment, and the face barely visible;
(3) Character diversity: photorealistic human and anthropomorphic cartoon is increasingly appearing in digital media.

\textbf{Why RealWorld-LipSync Matters.}
Our benchmark is a paradigm shift in evaluation. First, it reveals failure modes not seen in synthetic benchmarks, models that score well on HDTF or AIGC-LipSync often fail to work on ours due to texture mismatch, identity shift, motion artifacts etc under bad cases. Second, including corner cases such as stylized contents and extreme lightings to push forward the innovations to generalizable solutions instead of dataset-specific optimization. We will publicly release the benchmark to enable strict and repeatable research.

\subsection{Quantitative Evaluations}
We first evaluate on HDTF, a standard talking head benchmark commonly used in prior works. To provide a controlled test environment for our method to compete with the SOTA approaches, we take quantitative evaluations on this dataset on previous methods, including TalkLip~\cite{talklip}, IP-LAP~\cite{IP-LAP}, Diff2Lip~\cite{diff2lip}, MuseTalk~\cite{musetalk}, LatentSync~\cite{latentsync}, and OmniSync~\cite{omnisync}. And we took it on the RealWorld- Lipsync to test the real world cross domain ability of model.

Our evaluation uses multiple metrics to assess different aspects of performance. For visual quality and temporal continuity, we report FID~\cite{fid} and FVD~\cite{fvd}. CSIM~\cite{csim} assesses the degree of identity preservation. To capture image quality, we use no-reference metrics: NIQE~\cite{niqe}, BRISQUE~\cite{brisque}, and HyperIQA~\cite{hyperiqa}.LMD~\cite{lmd} for geometrical precision and LSE-C~\cite{wav2lip} for audio-visual alignment are used to measure the accuracy of lip synchronization. For the RealWorld-LipSync benchmark, we define the Generation Success Rate (GSR) to measure the percentage of videos that can be successfully generated. Evaluate only samples successfully generated by all the methods for a fair comparison of the quality metrics.

\begin{figure}[t]
    \centering
    \includegraphics[width=\columnwidth]{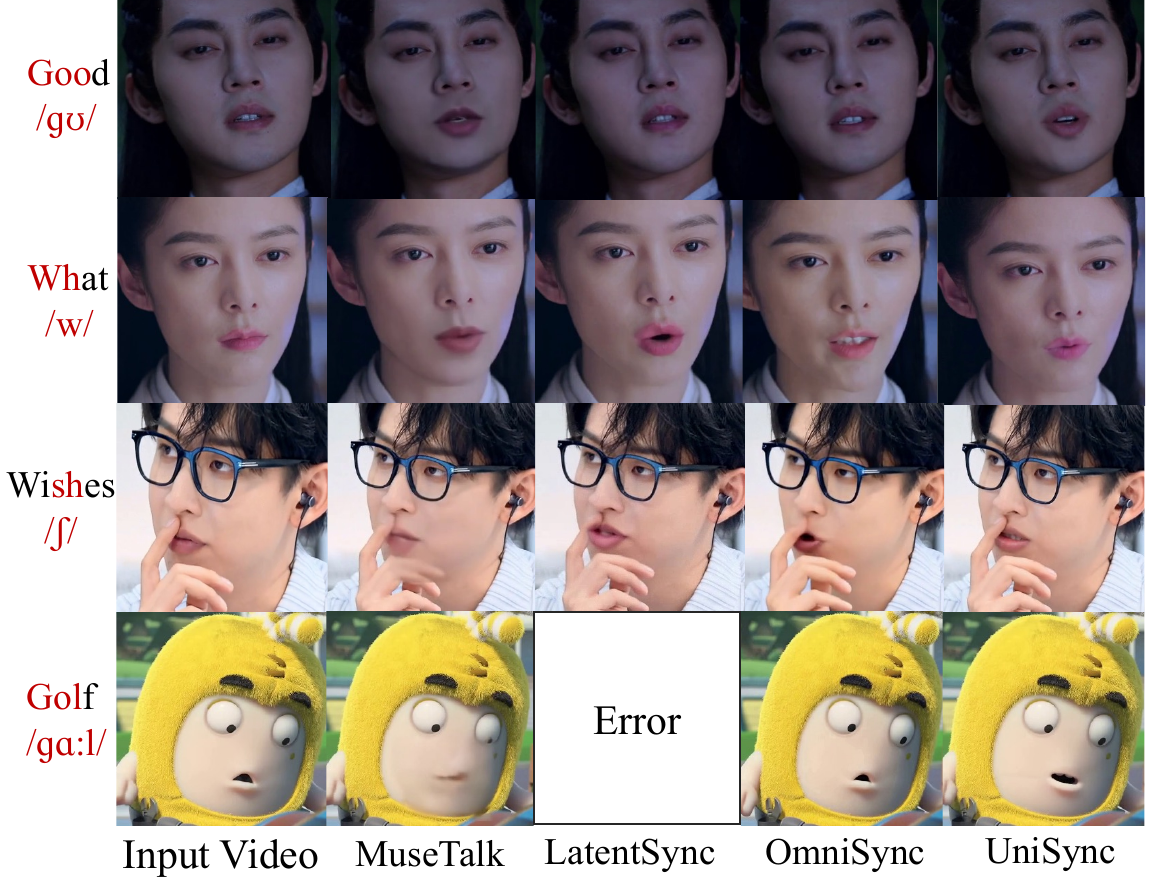}
    \caption{Quantitative evaluation results compared with recent SOTA methods MuseTalk, LatentSync, and OmniSync. Our method achieves the most accurate lip synchronization and demonstrates superior robustness across diverse real-world production scenarios.}
    \label{fig:evaluation}
\end{figure}

As shown in Table~\ref{tab:main_results}, our method achieves state-of-the-art results on the two benchmark datasets. On the HDTF dataset, UniSync has the best visual quality, the lowest FID and FVD, and is significantly better than the second best OmniSync. We also obtain the best identity preservation on CSIM and image quality on HyperIQA, proving the superiority of our mask-free pose-anchored training. More importantly, UniSync sets a new SOTA in terms of Generation Success Rate (GSR) on the difficult RealWorld-LipSync benchmark, which is more than 7\% ahead of the second-place method. This demonstrates the excellent robustness of our method to different real situations such as extreme light, stylized avatars and occlusion. Figure~\ref{fig:evaluation} shows qualitative comparisons with recently state-of-the-art methods. MuseTalk, LatentSync, and OmniSync are newer methods developed for high resolution cases and we will perform our visual comparisons among these. As for earlier methods such as TalkLip, IP-LAP, and Diff2Lip, they are limited to generate 256×256 resolution videos, which results in a lack of comparability with today’s modern production requirements. As shown in the figure, our method produces the most accurate lip synchronization and demonstrates superior visual quality across various challenging real-world scenarios.

\begin{table}[t]
    \centering
    \resizebox{\columnwidth}{!}{
    \setlength{\tabcolsep}{0.7pt}
    %\small
        \begin{tabular}{@{}lrrrrrrr@{}}
            \toprule
            \textbf{Methods} & \textbf{FID} $\downarrow$ & \textbf{FVD} $\downarrow$ & \textbf{CSIM} $\uparrow$ & \textbf{NIQE} $\downarrow$ & \textbf{BRISQUE} $\downarrow$ & \textbf{HyperIQA} $\uparrow$ & \textbf{LSE-C} $\uparrow$\\
            \toprule
            \textbf{w/o PAFS} & \cellcolor{lightblue}13.521 & 177.128 & 0.824 & \cellcolor{lightred}\bf{7.813} & \cellcolor{lightred}\bf{52.277} & \cellcolor{lightred}\bf{46.897} & 4.812 \\
            \textbf{w/o TALI} & 13.742 & \cellcolor{lightblue}167.277 & \cellcolor{lightblue}0.830 & 8.438 & 57.123 & 41.258 & \cellcolor{lightblue}4.855 \\
            \textbf{w/o Gaussian} & 14.841 & 191.128 & 0.805 & 8.852 & 58.441 & 40.019 & 4.390 \\
            \textbf{UniSync} & \cellcolor{lightred}\bf{12.976} & \cellcolor{lightred}\bf{141.573} & \cellcolor{lightred}\bf{0.869} & \cellcolor{lightblue}8.023 & \cellcolor{lightblue}55.988 & \cellcolor{lightblue}45.818 & \cellcolor{lightred}\bf{5.091} \\
            \bottomrule
        \end{tabular}
    }
    \caption{Ablation study for the proposed UniSync.}
    \label{tab:ablation}
\end{table}

\begin{table}[t]
    \centering
    \resizebox{\columnwidth}{!}{
    \setlength{\tabcolsep}{3pt}
    \begin{tabular}{@{}crrrrrrr@{}}
        \toprule
        $\tau_{inj}$ & \textbf{FID}$\downarrow$ & \textbf{FVD}$\downarrow$ & \textbf{CSIM}$\uparrow$ & \textbf{NIQE}$\downarrow$ & \textbf{BRISQUE}$\downarrow$ & \textbf{HyperIQA}$\uparrow$ & \textbf{LSE-C}$\uparrow$ \\
        \toprule
        \bf{0.0} & 13.742 & 167.277 & 0.830 & 8.438 & 57.123 & 41.258 & 4.855\\
        \bf{0.2} & 14.219 & 179.145 & \cellcolor{lightblue}0.915 & 8.154 & \cellcolor{lightblue}53.505 & 44.782 & 4.652\\
        \bf{0.4} & \cellcolor{lightblue}13.006 & 154.879 & 0.694 & \cellcolor{lightred}\bf{7.975} & \cellcolor{lightred}\bf{53.198} & 44.423 & 4.678\\
        \bf{0.6} & 14.316 & \cellcolor{lightblue}153.091 & 0.772 & 8.124 & 56.504 & 42.455 & 4.992\\
        \bf{0.8} & \cellcolor{lightred}\bf{12.976} & \cellcolor{lightred}\bf{141.573} & 0.869 & \cellcolor{lightblue}8.023 & 55.988 & \cellcolor{lightred}\bf{45.818} & \cellcolor{lightblue}5.091 \\
        \bf{1.0} & 13.538 & 160.894 & \cellcolor{lightred}\bf{0.921} & 8.465 & 56.773 &\cellcolor{lightblue}45.349 & \cellcolor{lightred}\bf{5.264} \\
        \bottomrule
    \end{tabular}
    }
    \caption{Sensitivity analysis of the injection ratio $\tau_{inj}$ in TALI. The optimal ratio is 0.8, achieving the best trade-off between temporal consistency (FVD) and generation quality.}
    \label{tab:refer_ratio}
\end{table}

\begin{figure}[t]
    \centering
    \includegraphics[width=\columnwidth]{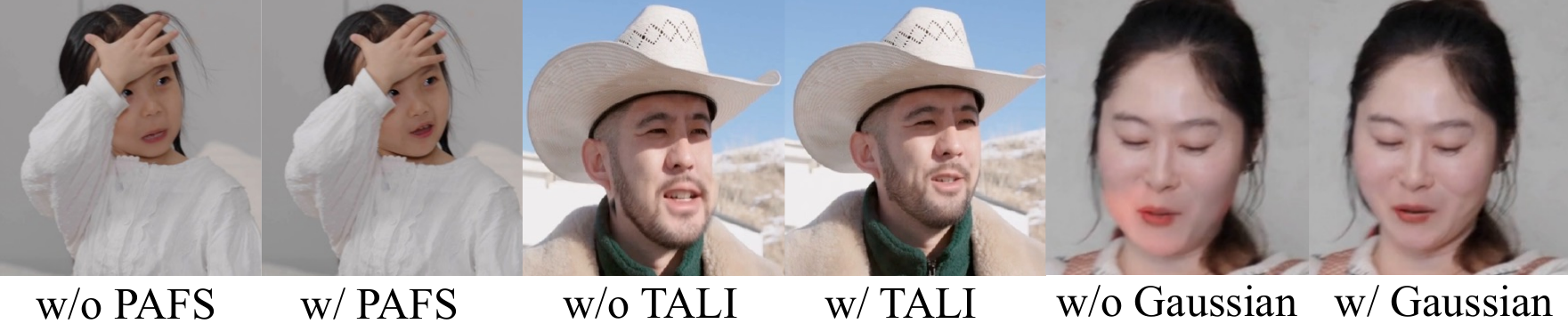}
    \caption{Ablation Study for PAFS, TALI and Gaussian smooth.}
    \label{fig:ablation}
\end{figure}

\subsection{Ablation Study}
Through a series of ablation studies, we find that the key parts of this paper, the Pose-Anchored Fidelity Strategy(PAFS), Temproal-Adaptive Latent Injection(TALI), Gaussian-Based Smooth Compositing Strategy, are all effective. From Table~\ref{tab:ablation} and Figure~\ref{fig:ablation}, we can conclude that all the ablations are important.

The removal of TALI is noticeable in the degradation of FVD and NIQE, confirming the importance of TALI in maintaining the background stability and temporal continuity. Excluding Gaussian smooth compositing (w/o Gaussian), the FID and FVD score is also higher, which shows that a seamless boundary compositing needs to be in order for a visual quality and temporal consistencies. Interestingly, the variant w/o PAFS obtains better scores on some of the perceptual metrics such as NIQE, BRISQUE and HyperIQA. This is due to the rich aesthetic prior of the pretrained AI2V backbone. But due to the lack of constraint, this is done at a high price: CSIM drops down to 0.824 and FVD gets worse. This indicates that without pose anchoring, the model hallucinates entirely new textures, and changes the identity of the subject, yielding very nice images that are not faithful to the original, thus failing at the most basic criterion of a video dubbing system. Therefore, our framework can achieve high-fidelity, stable structure and good temporal consistency at the same time.

For detail proof of TALI, we analyze the sensitivity of the injection ratio $\tau_{\text{inj}} \in [0,1]$ in TALI, which controls the strength of original video latent injection. Table~\ref{tab:refer_ratio} shows that $\tau_{\text{inj}}= 0.8$ obtain the best performance across most metrics. Lower values harm temporal consistency, but higher values over constrain generation though it yields better CSIM. Thus we validate that moderate latent injection achieve the trade-off of temporal cohesiveness and generative flexibility.

\begin{table}[t]
    \centering
    \resizebox{\columnwidth}{!}{
    \setlength{\tabcolsep}{3pt}
    \begin{tabular}{lcccc}
        \toprule
        \textbf{Metrics} & \textbf{MuseTalk} & \textbf{LatentSync} & \textbf{OmniSync} & \textbf{Ours} \\
        \midrule
        \textbf{Lip Sync Accuracy} & 2.44 & \cellcolor{lightred}{\bf{3.52}} & 3.37 & \cellcolor{lightblue}3.49 \\
        \textbf{Identity Preservation} & 2.30 & 3.12 & \cellcolor{lightblue}{3.62} & \cellcolor{lightred}\bf{3.83} \\
        \textbf{Timing Stability} & 2.28 & 3.14 & \cellcolor{lightblue}3.64 & \cellcolor{lightred}\bf{3.68} \\
        \textbf{Image Quality} & 2.08 & 3.07 & \cellcolor{lightblue}3.68 & \cellcolor{lightred}\bf{4.12} \\
        \textbf{Video Realism} & 2.14 & 2.99 & \cellcolor{lightblue}3.46 & \cellcolor{lightred}\bf{3.86} \\
        \bottomrule
    \end{tabular}
    }
    \caption{User study results comparing perceptual quality across different methods. Participants rate each metric on a scale of 1-5 (higher is better).}
    \label{tab:user_study}
\end{table}

\subsection{User Study}
In addition to quantitative data, we conducted a survey on 50 participants to study the five aspects of users' sense of the generated lip synchronization video, which are lip sync accuracy, identity consistency, timing stability, image quality, and realistic sense of the video. As shown in Table~\ref{tab:user_study}, our method has the best or comparable results in all dimensions. The worth mentioning is that we get the best result on timing stability and image quality as what we supposed that is to say our pose-anchor training strategy and blending can produce temporally orderly, pleasing result. The results are in line with the quantitive results and prove that UniSync has better perceptual quality.
%To complement quantitative metrics, we conduct a user study with 50 participants to evaluate perceptual quality across different dimensions. Participants rate videos on a 5-point Likert scale across five aspects: Lip Sync Accuracy, Identity Preservation, Timing Stability, Image Quality, and Video Realism. As shown in Table~\ref{tab:user_study}, our method achieves the best or comparable scores across all dimensions. Notably, we obtain the highest scores in Timing Stability and Image Quality, validating that our pose-anchored training and blending strategy produce temporally coherent and visually pleasing results. While OmniSync shows competitive performance in Image Quality, our method consistently outperforms it in critical aspects like Timing Stability and Lip Sync Accuracy. MuseTalk receives significantly lower ratings across all dimensions, particularly in Image Quality, indicating poor visual fidelity. These human evaluation results align with our quantitative findings and confirm that UniSync delivers superior perceptual quality in real-world scenarios.

\section{Conclusion}
In this paper, we proposed UniSync, a unified framework for the critical challenges of high-fidelity video dubbing in diverse real-world scenarios. By combining mask-free pose-anchored training with mask-based blending consistent inference, UniSync achieves significant improvements in visual quality, temporal consistency, and generation stability. Our compact yet diverse training data enables robust performance across complex conditions including extreme lighting, stylized avatars, and occlusions. The proposed RealWorld-LipSync benchmark evaluates models under authentic real-world requirements. A large number of experiments show that UniSync significantly outperforms the state-of-the-art, the generation success rate on the challenging scenarios where the existing methods often break down is above 93\%. We think our framework and benchmark set up a new bar for practical and high-quality video dubbing at scale.
\clearpage
\bibliographystyle{named}
\bibliography{ijcai26}

\end{document}